\documentclass{article}

    \PassOptionsToPackage{numbers, compress}{natbib}

\usepackage[preprint]{neurips_data_2022}
\usepackage{tabularx,booktabs}

\usepackage[utf8]{inputenc} %
\usepackage[T1]{fontenc}    %
\usepackage[pagebackref,breaklinks,colorlinks]{hyperref}
\usepackage{url}            %
\usepackage{booktabs}       %
\usepackage{amsfonts}       %
\usepackage{nicefrac}       %
\usepackage{microtype}      %
\usepackage{xcolor}         %

\usepackage{dsfont}
\usepackage{algorithm}
\usepackage{algorithmic}
\usepackage{subfigure}
\usepackage{graphicx}
\usepackage{epsfig}
\usepackage{amsmath}
\usepackage{wrapfig}
\usepackage{multirow}
\usepackage{amssymb}
\usepackage[capitalize,noabbrev]{cleveref}

\title{Benchmarking the Robustness of LiDAR-Camera Fusion for 3D Object Detection}

\def\sota{state-of-the-art}
\def\robustnuScenes{nuScenes-R }
\def\robustwaymo{Waymo-R}
\def\waymo{Waymo}
\def\nus{nuScenes}

\newcommand{\pp}{PointPillars}

\def\github{\href{https://github.com/kcyu2014/lidar-camera-robust-benchmark}{https://github.com/kcyu2014/lidar-camera-robust-benchmark}}

\newcommand{\mypara}[1]{\vspace{1mm}\noindent\textbf{#1}}

\author{
  Kaicheng Yu$^{1}$\footnotemark[1] \quad
  Tang Tao$^{2}$\footnotemark[1] \footnotemark[2] \quad
  Hongwei Xie$^{1}$\footnotemark[1] \quad 
  Zhiwei Lin$^{3}$  \quad
  Zhongwei Wu$^{1}$ \quad \\
  \textbf{Zhongyu Xia}$^{3}$ \quad
  \textbf{Tingting Liang}$^{3}$\quad  %
  \textbf{Haiyang Sun}$^{1}$ \quad
  \textbf{Jiong Deng}$^{1}$ \quad \\
  \textbf{Dayang Hao}$^{1}$ \quad
  \textbf{Yongtao Wang}$^{3}$ \quad
  \textbf{Xiaodan Liang}$^{2}$ \quad
  \textbf{Bing Wang$^{1}$}\footnotemark[4] \quad
  \\
$^1$ Autonomous Driving Lab, Alibaba Group, China \\
$^2$ Shenzhen Campus, Sun Yat-sen University, China \\
$^3$ Wangxuan Institute of Computer Technology, Peking University, China\\
 \\
  \texttt{\{kaicheng.yu.yt,trent.tangtao,hongwei.xie.90,xdliang328,blucewang6\}@gmail.com} \\
  \texttt{\{zhiweilin,tingtingliang,wyt\}@pku.edu.cn}  \quad\\
}

\begin{document}

\maketitle
\renewcommand{\thefootnote}{\fnsymbol{footnote}}
\footnotetext[1]{Equal Contribution.}
\footnotetext[2]{Work done during an internship at DAMO Academy, Alibaba Group.}
\footnotetext[4]{Corresponding Author.}
\renewcommand{\thefootnote}{\arabic{footnote}}

\begin{abstract}
\label{subsec:abstract}
There are two critical sensors for 3D perception in autonomous driving, the camera and the LiDAR. The camera provides rich semantic information such as color, texture, and the LiDAR reflects the 3D shape and locations of surrounding objects.
People discover that fusing these two modalities can significantly boost the performance of 3D perception models as each modality has complementary information to the other. 
However, we observe that current datasets are captured from expensive vehicles that are explicitly designed for data collection purposes, and cannot truly reflect the realistic data distribution due to various reasons.
To this end, we collect a series of real-world cases with noisy data distribution, and systematically formulate a robustness benchmark toolkit, that simulates these cases on any clean autonomous driving datasets. 
We showcase the effectiveness of our toolkit by establishing the robustness benchmark on two widely-adopted autonomous driving datasets, \nus{} and \waymo{}, then, to the best of our knowledge, holistically benchmark the \sota{} fusion methods for the first time. We observe that: i) 
most fusion methods, when solely developed on these data, tend to fail inevitably when there is a disruption to the LiDAR input; 
ii) the improvement of the camera input is significantly inferior to the LiDAR one. We further propose an efficient robust training strategy to improve the robustness of the current fusion method. The benchmark and code are available at \github{}.

\end{abstract}

\section{Introduction}

\begin{figure}[!t]
	\centering
	\vspace{-0.2cm}
    \includegraphics[width=1.\textwidth]{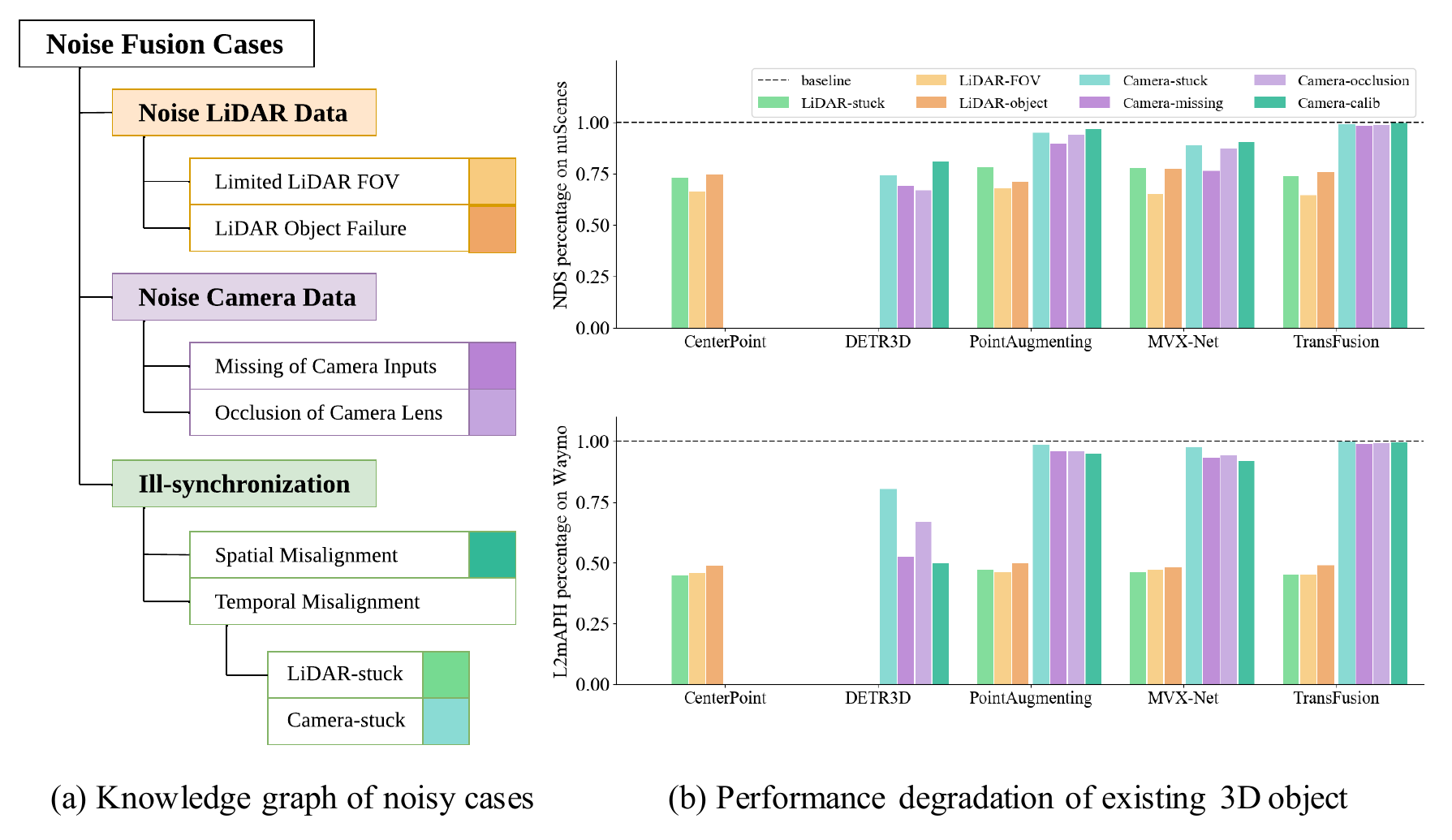}
    \vspace{-0.3cm}
    \caption{\textbf{Benchmarking the robustness of state-of-the-art 3D detection methods.} \textbf{(a)} We provide a knowledge graph of all noisy fusion cases. \textbf{(b)} We report the performance of current methods on two robustness datasets, \waymo{}-R and \nus{}-R, which are generated by our proposed toolkit. We observe that current fusion methods are more robust to image disruption rather than the LiDAR one. 
    }
	\vspace{-5mm}
	\label{fig:performance_degradation}
\end{figure}

3D detection has received extensive attention as one of the fundamental tasks in autonomous driving scenarios~\cite{Yan2018SECONDSE,Shi2019PointRCNN3O,Yang20203DSSDP3,Shi2020PVRCNNPF,lang2019pointpillars,abs-2203-17270,reading2021categorical,huang2021bevdet,abs-2204-05088,park2021pseudo}. Recently, fusing the two common modalities, input from the camera and LiDAR sensors, has become a de-facto standard in the 3D detection domain as each modality has complementary information of the other~\cite{chen2017mv3d,vora2020pointpainting, wang2021pointaugmenting, sindagi2019mvxnet, xu2021fusionpainting,Huang2020EPNetEP,Yin2021MVP}. 
Similar to other literature in the computer vision community, a common approach to showcase the effectiveness of a proposed fusion method is to validate it on the existing benchmark datasets~\cite{sun2020waymo,caesar2020nuscenes}, which are usually collected from explicitly designed, expensive data collection vehicles to minimize any potential error from the hardware setup.

However, we discover that the data distribution of these popular datasets can be drastically different from the realistic driving scenarios due to various reasons: i) there can be uncontrollable external reasons, such as splatted dirt or BIOS malfunctions, that temporarily disable the input of certain sensors; ii) the inputs can be difficult to synchronize due to external and internal reasons, like bumpy road or system clock misalignment. 
Therefore, we cannot estimate the performance of methods under these realistic settings, and evaluating only on these clean datasets is not trustworthy when deployed to realistic scenarios.

To this end, we close this research gap by proposing a novel toolkit that transforms any clean benchmark dataset, which has the camera and LiDAR input modality, into a robustness benchmark to simulate realistic scenarios. We first conduct a systematic overview of potential sensor noisy cases, both for the camera and LiDAR, based on realistic driving data, and summarized a knowledge graph in Figure~\ref{fig:performance_degradation}(a). Specifically, we identify seven unique cases under three categories, two for noisy LiDAR cases, two for noisy camera cases, and three for ill-synchronization cases.
We then carefully study each case and construct a code toolkit to transform the clean data into a realistic data distribution. 

To verify the effectiveness of our approach, we apply our toolkit to two large-scale popular benchmark datasets for autonomous driving, \nus{} and \waymo{}. 
Note that, though these noisy cases rarely appear in realistic scenarios, we convert all data from a dataset to fully explore the robustness of a given method in an extreme manner.
We collect two single modalities and three fusion state-of-the-art methods and benchmark them on the generated datasets. In Figure~\ref{fig:performance_degradation}~(b), we observe several surprising findings: i) state-of-the-art fusion methods tend to fail inevitably when the LiDAR sensor encounters failures due to their fusion mechanism heavily relies on the LiDAR input; ii) fusing the camera input only brings a marginal improvement, suggesting either the current methods fail to sufficiently leverage the information from the camera or the camera information did not carry the complementary information as intuited. Note that the toolkit only generates one failure case at a time, we do not create a robust benchmark that has multiple malfunctions at the same time. Though the main purpose of this work is to create a robustness bnechmark, we nonetheless provide a simple method, which finetunes the model on these robustness scenarios, and show that it moderately improves the robustness of current methods. However, there is still a large performance gap when compared to the results of the clean settings. 

We argue that an ideal goal of a fusion framework is when there is only a single modality sensor failure, the performance should not be worse than the method works on the other modality. Otherwise, the current fusion methods should be replaced by using two separate networks for each modality and performing fusion by post-processing steps~\cite{chen2017mv3d, pang2020clocs}. 
We hope our work can shed light on developing robust fusion method that can be truly deployed to the autonomous vehicle.

In summary, our main contribution can be summarized as follow: i) we systematically study the noisy sensor data in the realistic driving scenarios and propose a novel toolkit that can transform any autonomous driving benchmark datasets, that contain camera and LiDAR input, into a robustness benchmark; ii) to the best of our knowledge, we are the first to benchmark existing methods under the same settings and find that current fusion methods has a fundamental flaw and can fail inevitably when there is a LiDAR malfunction.

\section{Related Work}
In this section, we provide a literature review of current fusion methods in the 3D detection domain and the robustness evaluation. 

\mypara{Fusion methods in 3D detection.}
LiDAR and camera are two types of complementary sensors for 3D object detection in autonomous driving. 
In essence, the LiDAR sensor provides an accurate depth and shape information of the surrounding world in form of sparse point clouds \cite{qi2017pointnet++,Qi2017PointNetDL,qi2018frustumPF,Shi2019PointRCNN3O,Yang20203DSSDP3,li2021lidar,Zhou2018VoxelNetEL,Wang2020PillarbasedOD, yin2021center,Shi2020PVRCNNPF}, while the camera sensor provides an RGB-based image that contains rich semantic and texture information \cite{lu2021geometry, Roddick2019OrthographicFT, liu2021autoshape, kumar2021groomed, zhang2021objects, zhou2021monocular, reading2021categorical, wang2021progressive, wang2021depth,philion2020lift,reading2021categorical, huang2021bevdet,abs-2204-05088}. 
Recently, fusing these modalities to leverage the complementary information becomes a de-facto standard in the 3D detection domain. 
Based on the fusion mechanism location, these methods can be divided into three categories, early, deep, and late fusion schemes. 

Early fusion methods mainly concatenate the image features to the original LiDAR point to enhance the representation power. Specifically, these methods rely on the LiDAR-to-world and camera-to-world calibration matrix to project a LiDAR point on the image plane, where it serves as a query of image features~\cite{vora2020pointpainting, wang2021pointaugmenting, sindagi2019mvxnet, xu2021fusionpainting,Huang2020EPNetEP,Yin2021MVP}.  
Deep fusion methods extract deep features from some pre-trained neural networks for both modalities under a unified space~\cite{bai2022transfusion,li2022deepfusion,ku2018joint,chen2017mv3d,yoo20203dcvf, Liang2018DeepCF, Liang2019MultiTaskMF}, where a popular choice of such space is the bird's eye view~(BEV)~\cite{bai2022transfusion, yoo20203dcvf}.   
While both early and deep fusion mechanisms usually occur within a neural network pipeline, the late fusion scheme usually contains two independent perception models to generate 3D bounding box predictions for both modalities, then fuse these predictions using post-processing techniques~\cite{chen2017mv3d, pang2020clocs}. One benefit of these works is their robustness against single modality input failure. However, it is difficult to jointly optimize this line of methods due to the post-processing technique being usually non-differentiable. In addition, this pipeline has a potential higher deployment cost as it has three independent modules to be maintained.

\mypara{Robustness of LiDAR-Camera Fusion.}
In the domain of autonomous driving, there lacks such a benchmark dataset for robustness analysis to the best of our knowledge. 
There only exists a few preliminary attempts to investigate the robustness issue of the fusion methods. 
TransFusion \cite{bai2022transfusion} evaluates the robustness of different fusion strategies under three scenarios: splitting validation set into daytime and nighttime, randomly dropping images for each frame, misaligning LiDAR and camera calibration by randomly adding a translation offset to the transformation matrix from camera to LiDAR sensor.  However, TransFusion \cite{bai2022transfusion} mainly explores the robustness against camera inputs, and ignores the noisy LiDAR and temporal misalignment cases. DeepFusion~\cite{li2022deepfusion} examines the model robustness by adding noise to LiDAR reflections and camera pixels. Though the noise settings of DeepFusion~\cite{li2022deepfusion} are straightforward and brief, the noisy cases almost never appear in real scenes.

By contrast, we systematically review the autonomous driving perception system and identify three categories, in a total of seven cases of robustness scenarios, and propose a toolkit that can transform an existing dataset into a robustness benchmark.
We hope our work can help future research to benchmark the robustness of their methods fairly, and give researchers more insights about designing a more robust fusion framework. An ideal fusion framework should work better than a single modality, and will not be worse than the single modality model while the other modality fails. We hope the deep fusion method is better than late fusion methods that use post-processing techniques.

\section{Robust Fusion Benchmark}

In this section, we first provide a systematic overview of current autonomous driving vehicle systems with LiDAR and camera sensors to show why the data distribution of clean datasets can differ from real-world scenarios. These noisy data cases can be categorized into three broad classes, noisy LiDAR, noisy camera, and ill-synchronization cases. Then, we present a toolkit that can transform current clean datasets into realistic scenarios. 

\subsection{An overview of modern autonomous driving vehicle system}

\begin{figure}[h]
	\centering
    \includegraphics[width=1.\textwidth]{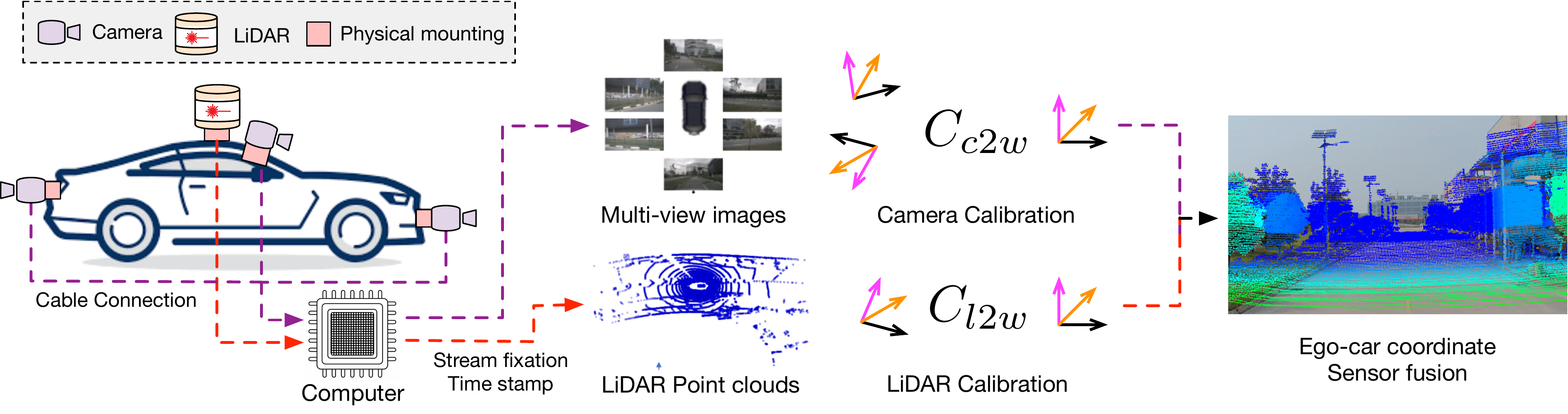}
    \vspace{-0.4cm}
    \caption{\textbf{Autonomous driving perception system with camera and LiDAR sensors.}}
	\label{fig:system}
\end{figure}

In \cref{fig:system}, we visualize a common design of the autonomous driving perception system, whose main components include the camera and LiDAR sensor, and an on-device computer. Specifically, the camera and LiDAR sensors are physically mounted on certain fixed locations of the vehicle and are connected to the computer via certain cables with communication protocols. In essence, the computer can access the data stream from the sensors and capture the data into a point cloud or image with a certain timestamp. As the raw data are in the sensor coordinate system, sensor calibration plays a major role in performing efficient coordinate transformation such that the perception system can recognize objects with respect to the ego-car coordinate system.
Based on our experience, each step of the aforementioned system can encounter certain failures or disruptions, and yield noisy data that are drastically different from the normal clean data. We identify three categories of cases and briefly discuss the potential reasons and consequences in Table~\ref{tab:noise-reason}, and provide a detailed case analysis later.

\begin{table}[t]
 \scriptsize \centering
 \addtolength{\tabcolsep}{8pt}
  \caption{\textbf{Common reasons for noisy data cases}. Based on the realistic experiences, we report various reasons that cause the noisy data cases. Note that both LiDAR and camera modality can encounter the temporal misalignment issue. }
  \label{tab:noise-reason}
  \centering
  \begin{tabularx}{0.9\textwidth}{c|c|c}
    \toprule
    Group & Reason & Consequent Case \\  
    \midrule
    \multirow{5 }{*}{Noisy LiDAR} & LiDAR sensor damage & no LiDAR point inputs \\
    \cmidrule(r){2-3}
    & Installation limitation of LiDAR & limited LiDAR field-of-view \\
    \cmidrule(r){2-3}
    & Low reflection rate of objects & LiDAR object failure \\
    \midrule
    \multirow{2}{*}{Noisy camera} & Camera sensors damage & Missing of corresponding image inputs \\
    \cmidrule(r){2-3}
    & Camera lens occlusion & Lens Occlusion  \\
    \midrule
    \multirow{7}{*}{Ill-synchronization} & Vibration during driving & Spatial misalignment \\
    \cmidrule(r){2-3}
    & Loose physical mounting & Spatial misalignment \\
    \cmidrule(r){2-3}
    & Instability of car computer & Temporal misalignment \\
    \cmidrule(r){2-3}
    & Temporary insufficient cable bandwidth & Temporal misalignment \\
    \cmidrule(r){2-3}
    & Sensor connection failure & Temporal misalignment \\
    \bottomrule
   \end{tabularx}
\end{table}

\subsection{Case analysis}
\label{subsec:benchmark}
In this section, we analyze the collected real-world noisy data cases of autonomous driving in detail. 

\subsubsection{Noisy LiDAR Data}
We identify two common cases that can cause noisy LiDAR data in practice.

\begin{figure}[t]
	\centering
    \includegraphics[width=1\textwidth]{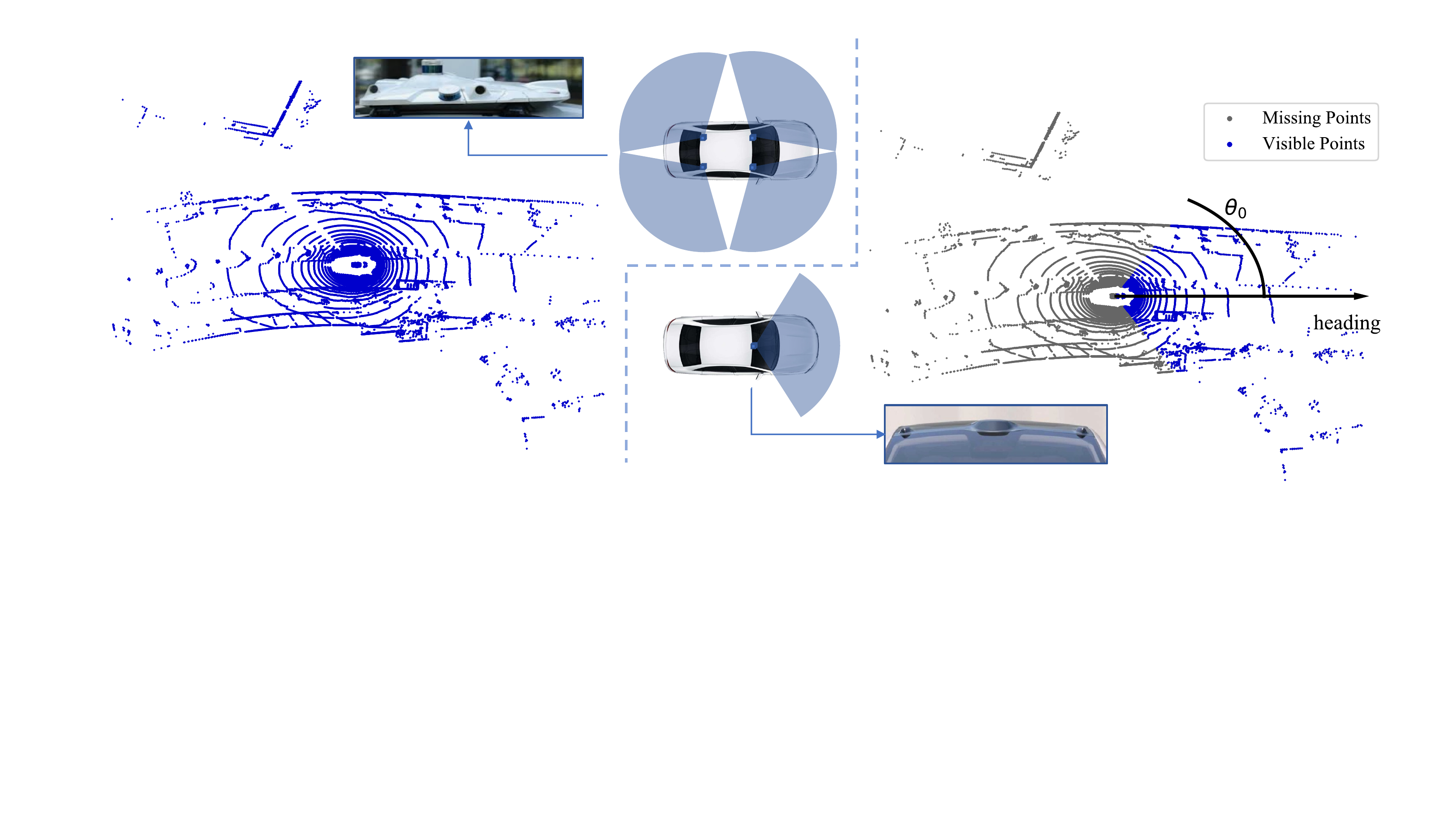}
    \caption{\textbf{Limited LiDAR field-of-view}. This figure is better viewed in color. (Left) We visualize the complete LiDAR point clouds that come from the data collection vehicle which has a complete sensor rack. (Right) 
    In realistic scenarios, the LiDAR is installed in a front-facing manner, yielding a limited FOV.  
    }
	\label{fig:lidar_drop_angle}
\end{figure}
\mypara{Limited LiDAR field-of-view~(FOV).} While most companies collect the LiDAR data whose field-of-view is 360 degrees, certain LiDAR data might not always be available for various reasons. For example, a certain type of vehicle only installs a front-facing semi-solid LiDAR sensor on the roof of the car instead of using a full rack, as shown in the right part of Figure~\ref{fig:lidar_drop_angle}. Another common reason might be the temporary occlusion of the LiDAR sensor. Without loss of generality, we first convert the coordinate of LiDAR points from Euclidean~(in $x,y,z$)  to polar coordinate system~($r, \theta, z$). We then can simulate such limited FOV by dropping the points that satisfy $\theta \in (-\theta_0, \theta_0)$. In practice, we set $\theta_0$ to 0, 60, and 90 degrees to simulate three commonly seen scenarios. 

\begin{figure}[t]
	\centering
    \includegraphics[width=0.9\textwidth, height=0.33\linewidth]{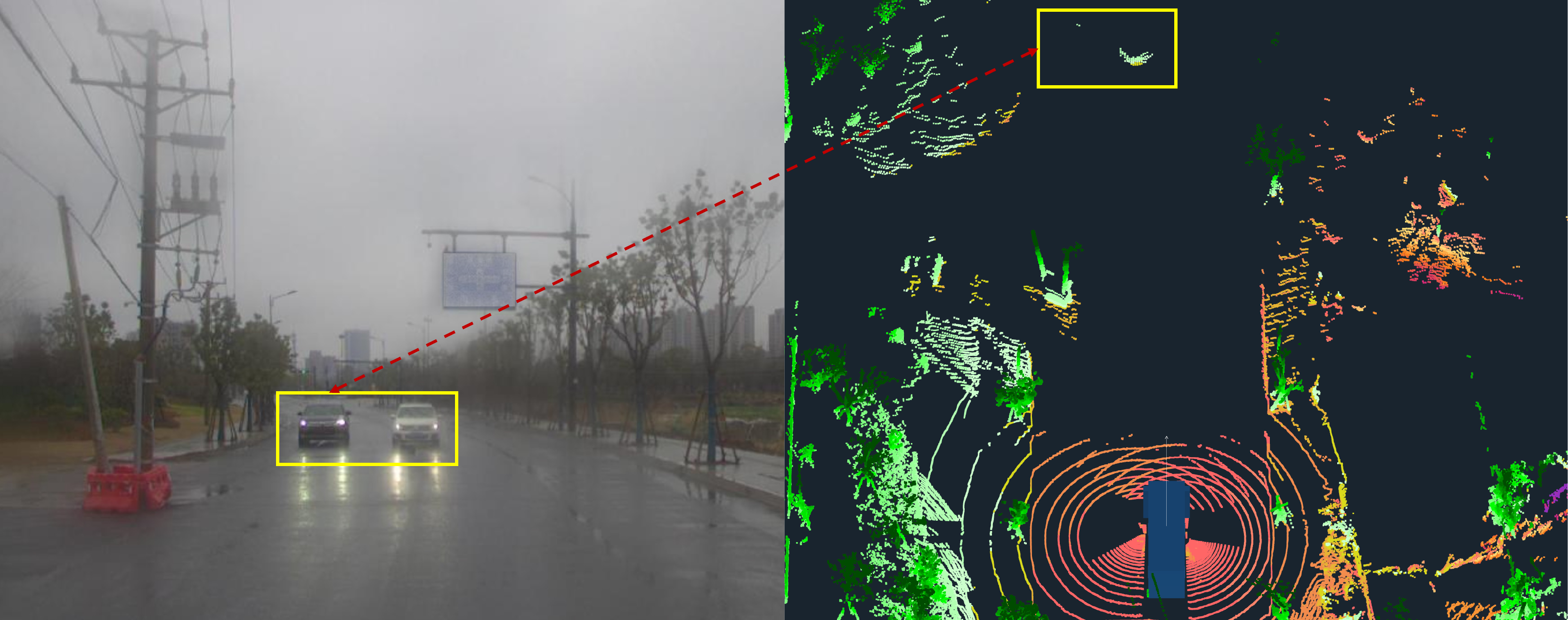}
    \caption{\textbf{LiDAR object failure}. On rainy days, the reflection rate of some common objects (e.g., the black car in the yellow box) is below the threshold of LiDAR hence causing the issue of object failure. This figure is better viewed in color. 
    }
	\label{fig:object_failure}
\end{figure}

\mypara{LiDAR object failure.}
One common scenario that people tend to overlook is that the LiDAR can be blind to objects under certain constraints. We show one example from the realistic data captured on a commercialized autonomous driving system in Figure~\ref{fig:object_failure}. We observe that the LiDAR point clouds are drastically different from two side-by-side cars, where the black car has nearly zero points while the white car has a normal point distribution. We dub this phenomenon LiDAR object failure. Without loss of generality, we simulate such scenarios by randomly dropping the points within a bounding box with a probability of 0.5. Note that we do not alter the camera input because the purpose is to benchmark the single modality input data.

\subsubsection{Noisy Camera Sensor} 

Different from the LiDAR module, the camera module is usually installed on much lower locations of the autonomous driving vehicles to cover the blind region of the LiDAR sensor. Such blind region is due to the fact that the LiDAR is usually installed on the roof of the car to maximize the visualization distance, while it cannot see the near-car region due to blockage. As such, the camera can be easily affected by the surrounding environment such as temporary generic object coverage or lens occlusion of dirt. We discuss these two scenarios in detail.

\mypara{Missing camera inputs.}
As the camera module is usually much smaller (within one centimeter) than the LiDAR sensor, the probability of covering half of the camera sensor is minimal. Thus, we drop the entire camera input to simulate such covering scenarios. In practice, we design two finer cases to perform a robust benchmark, dropping one camera at a time, and dropping all other cameras except the front one.

\begin{figure}[t]
	\centering
	\begin{minipage}{0.26\linewidth}
        \centerline{\includegraphics[width=\textwidth]{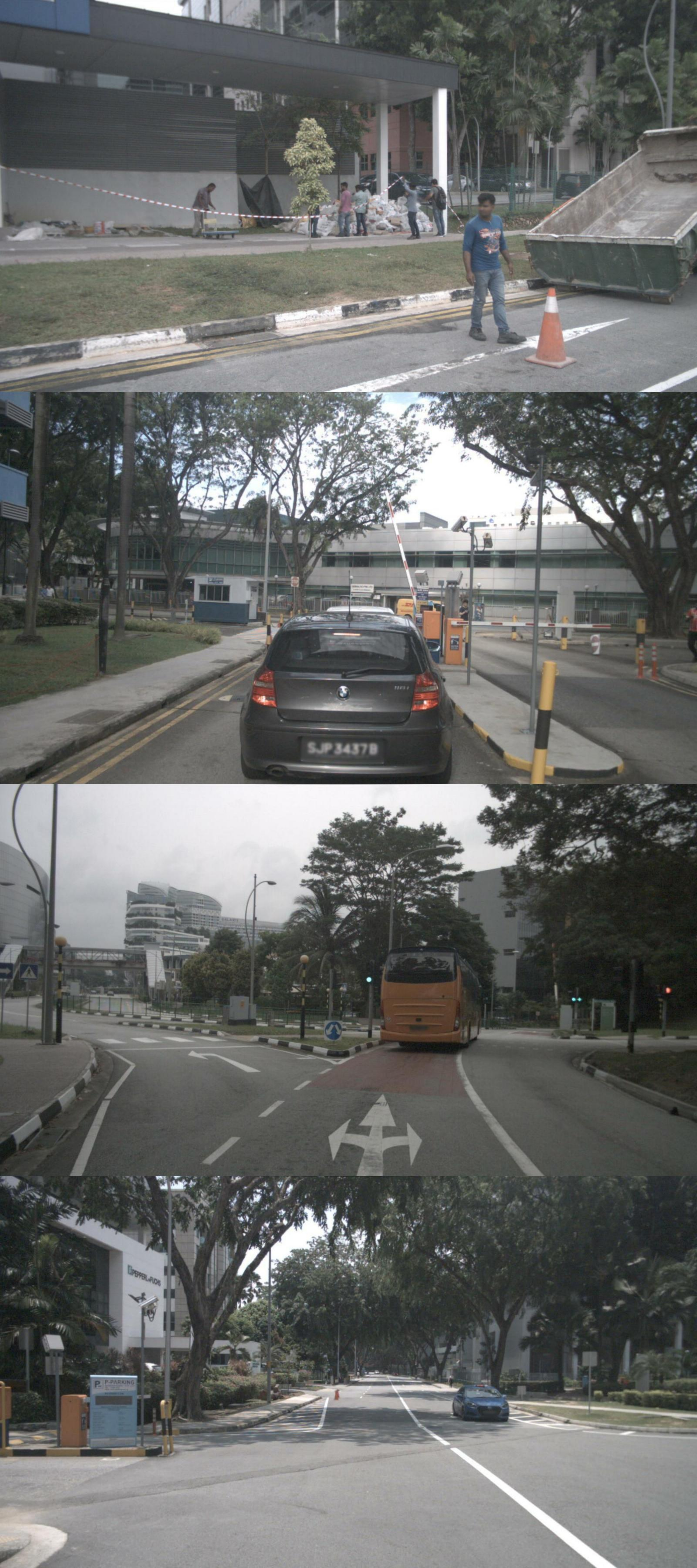}}
    \end{minipage}
    \begin{minipage}{0.26\linewidth}
        \centerline{\includegraphics[width=\textwidth]{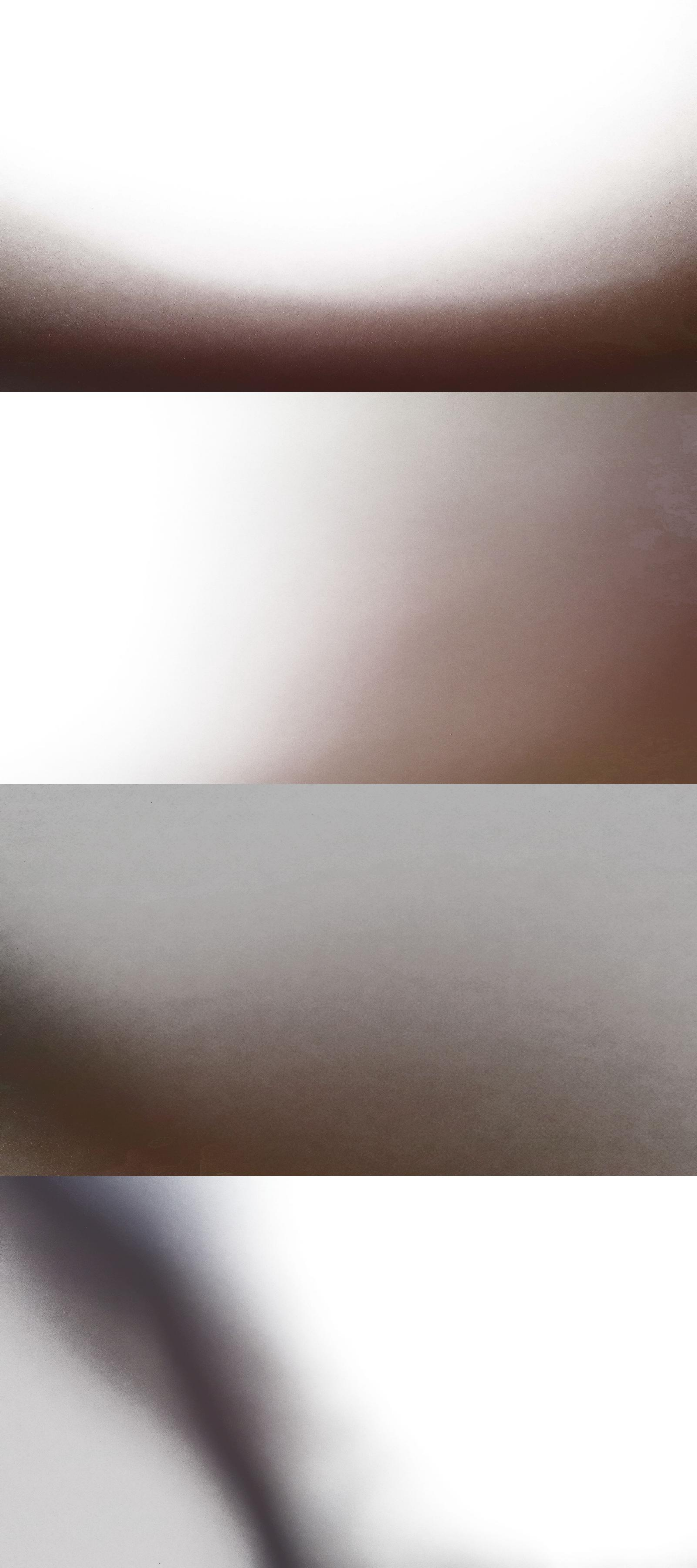}}
    \end{minipage}
    \begin{minipage}{0.26\linewidth}
        \centerline{\includegraphics[width=\textwidth]{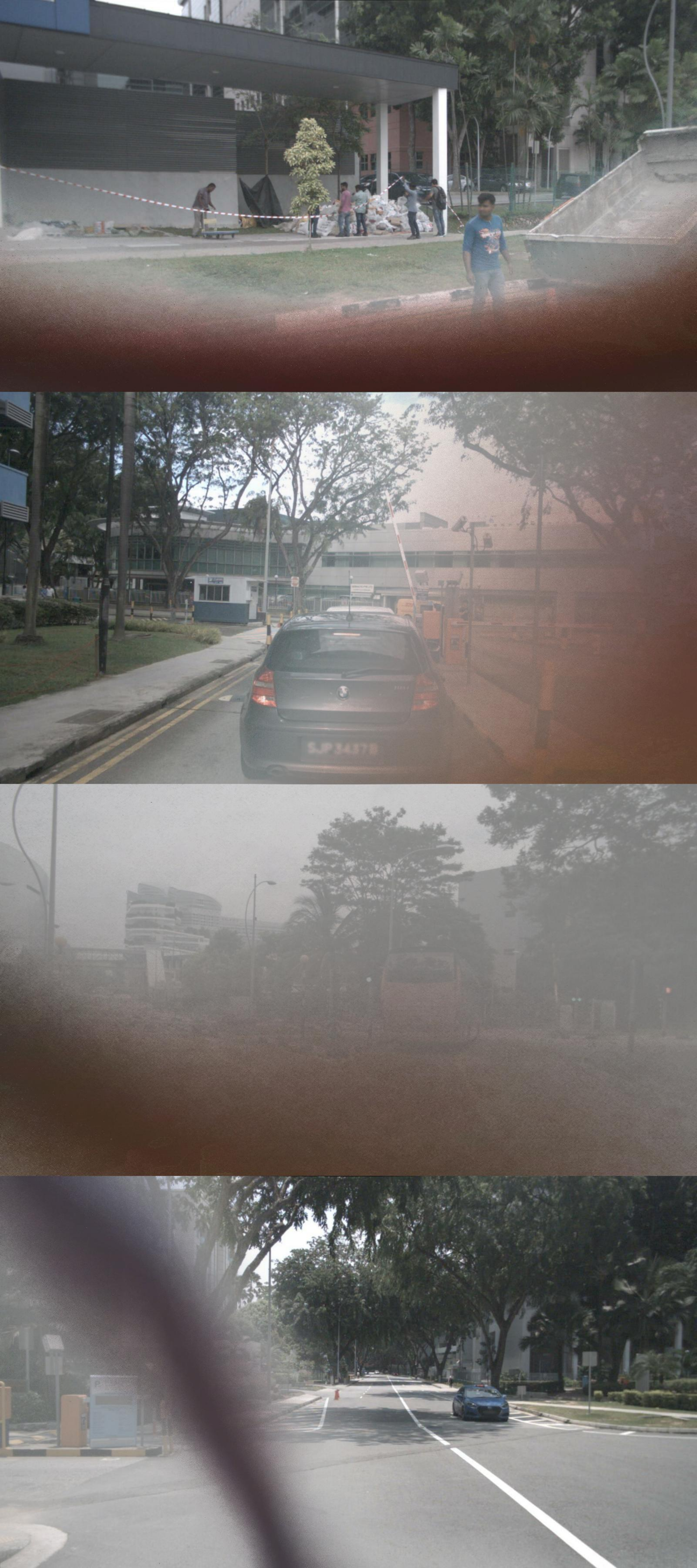}}
    \end{minipage}
    \caption{\textbf{Visualization of camera occlusion}. 
     We display the original images from four different scenarios in the \nus{} dataset (Left), the randomly sampled dirt occlusion masks (Middle), and the final composed images that simulating the occlusion (Right). 
    }
	\label{fig:mask_img}
\end{figure}

\mypara{Camera lens occlusion.} 
Another commonly seen camera covering problem is lens occlusion caused by non-transparent liquid or dirt. 
To simulate the occlusion of camera lenses in real scenes, we spray mud dots on a transparent film and cover the dirty film on the camera lens to take photos on a white background.
Then, we adopt an image matting algorithm to cut out the background part in the images and separate the masks of mud spots.
Finally, the separated masks are pasted on the images of nuScene or Waymo to simulate the occlusion of their camera lens, as illustrated in the \cref{fig:mask_img}.
In addition, we spray mud dots of different sizes and randomly move and rotate the film to create masks with different occlusion areas and occlusion ranges to enhance the diversity of the mask.

\subsection{Ill-synchronization}

As illustrated in Figure~\ref{fig:system}, the data stream is firstly fixed into a data frame with a given timestamp when passed into the on-device computer, then one needs to perform the coordinate transform via the camera-to-world and LiDAR-to-world matrix that is obtained by the calibration process. However, this leads to two potential ill-synchronization issues, spatial misalignment due to the external reasons of calibration matrix and temporal misalignment for both LiDAR and camera data due to internal system reasons. 

\begin{figure}[t]
	\centering
    \includegraphics[width=0.9\textwidth]{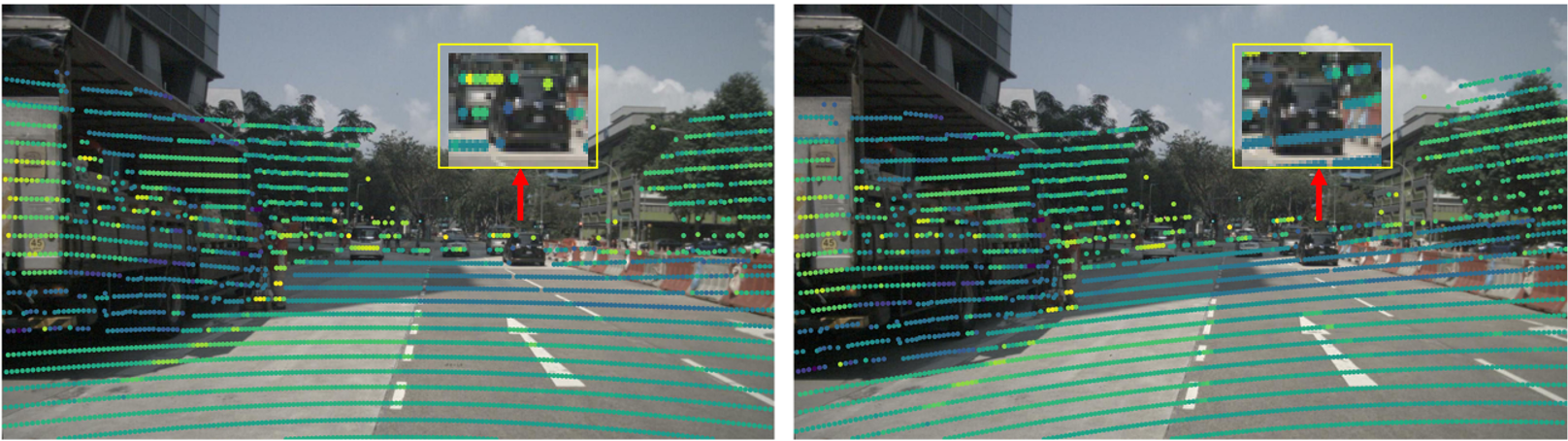}
    \caption{\textbf{Visualization of spatial misalignment effect}. As the physical size of a camera module is drastically smaller than the size of vehicle, the relative position of car center to the camera center will inevitably change due to various reasons, like the vibration during driving on the bumpy road, and since such noise happens all the time, it cannot be avoided using online calibration. We provide one visual example to showcase such misalignment. 
    }
	\label{fig:spatial_misalign}
\end{figure}

\mypara{Spatial misalignment.}
The bumping and shaking of vehicles lead to the disturbance of the extrinsic parameters of cameras and cause spatial misalignment between LiDAR and camera inputs. In addition, such errors can accumulate while the mileage of a vehicle is increasing. 
To simulate such a situation, we add random rotation and translation noise to the calibration of each camera independently. 
The range of noise rotation angle is from $0^{\circ}$ to $5^{\circ}$ and the translation range is from $1$ cm to $5$ cm to accord with the noise range in the real scene. Sensor calibration misalignment will cause spatial misalignment between point cloud and image, as shown in \cref{fig:spatial_misalign}.

\begin{figure}[t]
	\centering
    \includegraphics[width=0.9\textwidth]{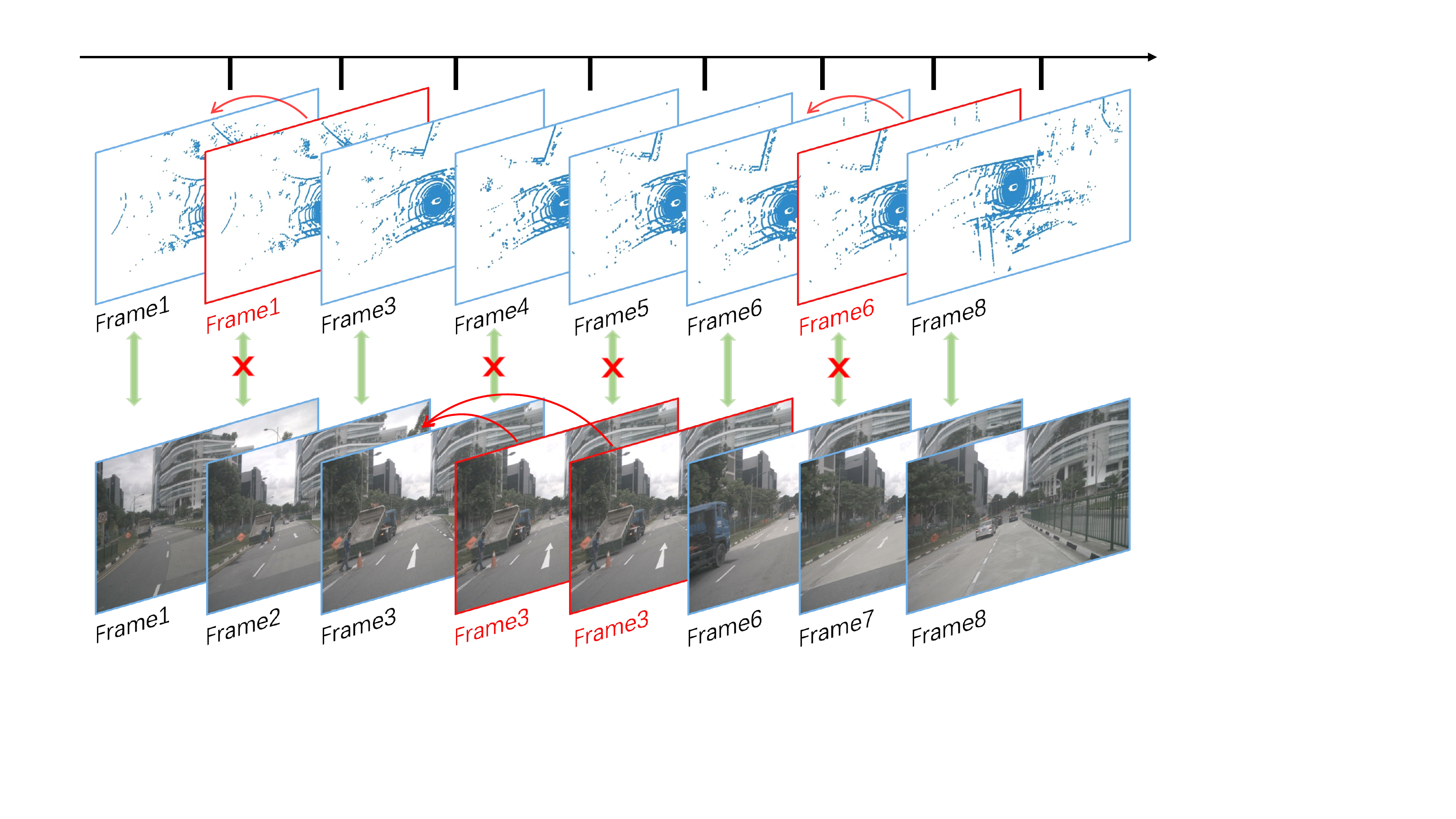}
    \caption{\textbf{Visualization of temporal misalignment}. Here we show one concrete example of temporal misalignment. This happens quite commonly on realistic autonomous vehicle. As the streaming data is first fixed with a certain time stamp then passed into the corresponding code module of deep learning model via system sockets, the timestamp of both modality sensors might not be always synchronized. To simulate such effect, we let the frame remains the same as previous frame when the data is ill-synchronized. We dub the phenomenon data stuck.  We consider two type of temporal misalignment, LiDAR stuck on the top of this figure and camera stuck on the bottom. 
    }
	\label{fig:temporal_misalign}
\end{figure}

\mypara{Temporal misalignment.} 
In a realistic autonomous driving system, failure of the system components is quite common throughout the time. One obvious consequence is the time-stamp of two modalities might not always be aligned. In some rare cases, the data frame of one modality can be stuck by over one minute depending on different system implementations. Here, we simulate such temporal misalignment in detail and provide one illustration in Figure~\ref{fig:temporal_misalign}.  
Initially, we apply nine levels of severity according to the percentage of stuck frames in all frames. The results show that the performance degradation of the 3D detection methods is linearly related to the percentage of stuck frames, as discussed in \cref{subsec:robustness analysis}. 
Thus, to reduce the load of the benchmarks, we only consider the case where the stuck frames are 50\% of all frames as the final benchmark setting.

\subsection{A toolkit to transform generic autonomous driving dataset into robustness benchmark}
\label{subsec:datasets}
To remove the randomness of benchmark comparison, we compose a toolkit that can transform any autonomous driving dataset into a robustness benchmark~\footnote{See our GitHub repository for more details. \github{}. }. In essence, we only simulate noisy data cases by altering the image and LiDAR data, the ground-truth annotation will remain the same as the 3D position of the object in the surrounding worlds will not change when the sensors malfunction. 

To facilitate future research, we leverage two popular large-scale autonomous driving datasets, \nus{} and \waymo{}, and benchmark state-of-the-art methods to evaluate their robustness for the first time to the best of our knowledge. We denote the newly created robustness benchmark \robustnuScenes{} and \robustwaymo{}.

\mypara{Evaluation Metrics.} 
To intuitively show the robustness of LiDAR-camera fusion methods, we simply use the performance and the relative performance degradation on our benchmark datasets as our evaluation metrics.
Specifically, the LiDAR-camera fusion model performance on the clean dataset is denoted as $P_{C}$ and its corresponding robustness performance against disruption type $d$ under severity level $l$ on the benchmark is denoted as $P_{R}^{d,l}$.
Then, we can estimate the model robustness $mP_{R}$ by averaging over all noise types and severity levels.
The formula can be summarized as follows:
\begin{equation}
    mP_{R}=\frac{1}{N_{d}}\sum\limits_{d=1}^{N_{d}}\frac{1}{N_{l}}\sum\limits_{l=1}^{N_{l}}P_{R}^{d,l},
\end{equation}
where 
$N_{d}$ is the number of disruption types and ${N_{l}}$ is the number of severity levels. 
The relative mean robustness performance of the model is defined as $R=mP_{R}/P_{C}$. 
The higher $R$ means the model is more robust to inferior senor fusion conditions. 
In practice, we adopt the mean Average Precision (mAP) and the weighted consolidated metric NDS as $P_{C}$ for \robustnuScenes{} and L2-mAP and L2-mAPH as $P_{C}$ for \robustwaymo{}.

\section{Benchmark Existing Methods}
We investigate and evaluate existing popular LiDAR-camera fusion methods with opening source code on our benchmark, including PointAugmenting~\cite{wang2021pointaugmenting}, MVX-Net~\cite{sindagi2019mvxnet}, and TransFusion~\cite{bai2022transfusion}. 
In addition, we also evaluate a LiDAR-only method, CenterPoint~\cite{nabati2021centerfusion}, and a camera-only method, DETR3D~\cite{wang2022detr3d}, for better comparison. 
It is worth noting that the metrics on waymo dataset focus on intersection of union~(IoU). However, strictly calculating the IoU of 3D bounding boxes is quite challenging for camera-based methods. Thus we reduce the IoU threshold to 0.3 and report the vehicle class for DETR3d on \waymo{} dateset.

\begin{table}[t]
  \tiny\centering\addtolength{\tabcolsep}{-2.7pt}
  \caption{\textbf{Benchmarking the robustness of state-of-the-art methods in all seven scenarios of the \robustnuScenes and \robustwaymo}. M denotes input modality, camera~(C) and LiDAR~(L), of state-of-the-art methods.
}
  \label{SOTA-table}
  \centering
 \begin{tabularx}{\textwidth}{l|c|ccc|ccc|cccc}
    \toprule
    \multicolumn{1}{c|}{\multirow{2.8}{*}{Approach}}&\multirow{2.8}{*}{M}&&&& \multicolumn{3}{c|}{LiDAR}  & \multicolumn{4}{c}{Camera} \\  
    \cmidrule(r){3-12}
     & & $P_{C}$ & $mP_{R}$ & $R$ & Stuck & FOV & Object & Stuck & Missing & Occlusion  & Calib 
  \\ \midrule
  \multicolumn{12}{c}{\robustnuScenes (mAP / NDS)}                 \\
    \midrule
     CenterPoint \cite{yin2021center} & L & 56.8 / 65.0 & 23.4 / 46.3 & 0.41 / 0.71 & 26.1 / 47.5 & 15.6 / 43.0 & 28.4 / 48.5 & - & - & - & -   \\
     DETR3D \cite{wang2022detr3d} & C & 34.9 / 43.4 & 17.6 / 31.5 & 0.50 / 0.73 & - & - & - & 17.3 / 32.3 & 14.5 / 29.9 & 14.3 / 29.0 & 24.2 / 35.0
  \\
     PointAugmenting \cite{wang2021pointaugmenting} & LC & 46.9 / 55.6 & 33.7 / 48.3 & 0.72 / 0.87 & 25.3 / 43.5 & 13.3 / 37.7 & 21.3 / 39.4 & 42.1 / 52.8 & 37.0 / 49.8 & 40.7 / 52.2 & 43.6 / 53.8
   \\
     MVX-Net \cite{sindagi2019mvxnet}& LC & 61.0 / 66.1 & 38.4 / 53.6 & 0.63 / 0.81 & 35.2 / 51.4 & 17.6 / 43.1 & 34.0 / 51.1& 48.3 / 58.8 & 32.7 / 50.6 & 45.5 / 57.6 & 50.8 / 59.9
 \\
     TransFusion \cite{bai2022transfusion}& LC & 66.9 / 70.9 & 52.8 / 63.1 & \textbf{0.79} / \textbf{0.89} & 33.4 / 52.3 & 20.3 / 45.8 & 34.6 / 53.6 & 65.9 / 70.2  & 64.9 / 69.7 & 65.5 / 70.0 &	66.5 / 70.7
   \\
     
    \midrule
    \multicolumn{12}{c}{\robustwaymo  (L2 mAP / L2 mAPH)}       \\

    \midrule
    CenterPoint \cite{yin2021center} & L & 66.0 / 63.4  & 30.6 / 29.4  & 0.46 / 0.46  & 29.5 / 28.3 & 30.3 / 29.1 & 32.1 / 30.9 & -  & - & - & - \\
     DETR3D \cite{wang2022detr3d} & C &16.2 / 15.7  & 10.1 / 9.8 & 0.62 / 0.62 & - & - & - & 13.0 / 12.6 & 8.4 / 8.2 & 10.9 / 10.5  & 8.0 / 7.8  \\
   PointAugmenting\cite{wang2021pointaugmenting} & LC & 52.5 / 50.7 & 39.6 / 38.3 & 0.75 / 0.76 &24.7 / 23.9 & 24.3 / 23.4 & 26.2 / 25.3 & 51.7 / 50.0 & 50.4 / 48.6 & 50.3 / 48.6 & 49.8 / 48.1  \\
     MVX-Net \cite{sindagi2019mvxnet}& LC & 59.7 / 54.1 & 44.3 / 40.1 & 0.74 / 0.74 & 27.5 / 24.9 &28.8 / 25.6 & 28.7 / 26.0 & 58.2 / 52.7 & 55.9 / 50.5 & 56.4 / 51.1 & 54.9 / 49.6  \\
     TransFusion \cite{bai2022transfusion}& LC &66.7 / 64.1 & 51.2 / 49.1  & 0.77 / 0.77 & 30.2 / 29.0 &30.2 / 29.0 & 32.7 / 31.3& 66.5 / 63.9 & 66.1 / 63.5 & 66.2 / 63.6 & 66.3 / 63.7   \\

    \bottomrule
    \multicolumn{12}{l}{Stuck: Temporal misalignment for both modalities. FOV: Limited LiDAR FOV. Object: LiDAR object failure. } \\ 
    \multicolumn{12}{l}{Missing: Missing camera inputs. Occlusion: Camera Lens Occlusion. Calib: Spatial misalignment of camera-to-world matrix. }
   \end{tabularx}
\end{table}

\begin{table}[t]
  \scriptsize \centering \addtolength{\tabcolsep}{1.3pt}
  \caption{\textbf{Robustness against LiDAR and camera modals of state-of-the-art architectures}. In short, the robust metric~(R) is computed by averaging the cases by the affecting modality.}
  \label{SOTA-table2}
  \centering
 \begin{tabularx}{0.85\textwidth}{l|c|c|cc|cc}
    \toprule
    \multicolumn{1}{c|}{\multirow{2.8}{*}{Approach}}& \multirow{2.8}{*}{Modality} & & \multicolumn{2}{c|}{LiDAR}  & \multicolumn{2}{c}{Camera} \\  
    \cmidrule(r){3-7}
 && $P_{C}$ & $mP_{R}$ & $R$ & $mP_{R}$ & $R$ \\
   \midrule
   \multicolumn{7}{c}{\robustnuScenes (mAP / NDS)}                 \\
    \midrule
     CenterPoint \cite{yin2021center}& L & 56.8 / 65.0 & 23.4 / 46.3 & 0.41 / 0.71  & - & - \\
     DETR3D \cite{wang2022detr3d}& C & 34.9 / 43.4 & - & - & 17.6 / 31.5 & 0.50 / 0.73   \\
     PointAugmenting \cite{wang2021pointaugmenting}&LC& 46.9 / 55.6 & 19.3 / 40.6 & 0.41 / \textbf{0.73}  & 40.9 / 52.2 &  0.87 / 0.94 \\
     MVX-Net \cite{sindagi2019mvxnet}&LC & 61.0 / 66.1 & 26.4 / 47.3 & \textbf{0.43} / 0.72 & 44.3 / 56.7 & 0.73 / 0.86 \\
     TransFusion \cite{bai2022transfusion}&LC & 66.9 / 70.9 & 26.9 / 49.1 & 0.40 / 0.69 & 65.7 / 70.1 & \textbf{0.98} / \textbf{0.99} \\
    \midrule
  \multicolumn{7}{c}{\robustwaymo  (L2 mAP / L2 mAPH)}       \\
    \midrule
    CenterPoint \cite{yin2021center}&L  & 66.0 / 63.4 & 30.6 / 29.4  & 0.46 / 0.46 & - & - \\
     DETR3D \cite{wang2022detr3d}&C &  16.2 / 15.7 & - & - & 10.1 / 9.8 & 0.62 / 0.62  \\
   PointAugmenting\cite{wang2021pointaugmenting}&LC  & 52.5 / 50.7 & 25.1 / 24.2 & 0.48 / 0.48 & 50.6 / 48.8 & 0.96 / 0.96 \\
     MVX-Net \cite{sindagi2019mvxnet}&LC & 59.7 / 54.1 & 28.3 / 25.5 & 0.47 / 0.47 & 56.4 / 51.0 &  0.94 / 0.94 \\
     TransFusion \cite{bai2022transfusion}&LC& 66.7 / 64.1 & 31.0 / 29.8 & 0.46 / 0.46 & 66.3 / 63.7 &  0.99 / 0.99\\

    \bottomrule
   \end{tabularx}
\end{table}

\subsection{Benchmark results}

The fusion robustness results are shown in \cref{SOTA-table}. 
Moreover, to analyze the robustness of models against LiDAR and camera disruptions, we present the $mP_{R}$ and $R$ of {LiDAR and camera modal} separately in \cref{SOTA-table2}.

In general, existing methods perform poorly on our robust fusion benchmark as shown in \cref{SOTA-table}, and there is vast room for improvement. Especially, for all LiDAR-camera fusion methods shown in \cref{SOTA-table2}, the LiDAR robustness is worse than camera robustness. 
Among the LiDAR-camera fusion methods we investigated, TransFusion achieves the overall best robustness. 
It is worth noting that the robustness against camera noise of TransFusion is unexpectedly outstanding. 
The average NDS of TransFusion only decreases from 70.9 to 70.1 on \robustnuScenes{} for camera failure and misalignment cases. 
However, the robustness against LiDAR noise of TransFusion is worse than other fusion methods.
We speculate that this is mainly due to the fact that camera information has a small effect on TransFusion.
There is only a slight improvement (1.5 NDS on nuScenes) when fusing camera information. Thus, when the camera information is missing or defective, the overall performance of TransFusion is not affected.
\par
When comparing the performance of LiDAR-camera fusion methods with single modality methods on our benchmark, we find all fusion methods have stronger robustness on both LiDAR and camera modality than single modality methods.  
This indicates that when encountering imperfect single modality inputs, the fusion methods can utilize another modality information to enhance the feature and predict the final outputs.

\begin{table}[t]
 \scriptsize\centering\addtolength{\tabcolsep}{-3.9pt}
  \caption{\textbf{Results of the limited LiDAR field-of-view case}. The angle ranges in brackets mean the visible angle range. ($-0$, $0$) means the extreme case when all LiDAR points are missing.}
  \label{tab:lidar-failure}
  \vspace{-0.2cm}
  \centering
  \begin{tabularx}{1\textwidth}{l|c|cccc|cccc}
    \toprule
    \multicolumn{1}{c|}{\multirow{2.8}{*}{Approach}} & \multirow{2.8}{*}{Modality} & \multicolumn{4}{c|}{\robustnuScenes (mAP / NDS)}  & \multicolumn{4}{c}{\robustwaymo  (L2 mAP / L2 mAPH)} \\  
    \cmidrule(r){3-10}
     && clean & ($-\pi/2$, $\pi/2$) &($-\pi/3$, $\pi/3$) & ($-0$, $0$) & clean & ($-\pi/2$, $\pi/2$) & ($-\pi/3$, $\pi/3$) & ($-0$, $0$) 
  \\
  
    \midrule
     CenterPoint \cite{yin2021center} & L &56.8 / 65.0 &23.5 / 47.7 &15.6 / 43.0 & 0 / 0 &66.0 / 63.4 &36.6 / 35.2 &30.3 / 29.1 &0 / 0   \\
     PointAugmenting \cite{wang2021pointaugmenting} & LC & 46.9 / 55.6 & 19.5 / 41.2 & 13.3 / 37.7 & 0 / 0 & 52.5 / 50.7  & 29.4 / 28.3  &  24.3 / 23.4  & 0 / 0   \\
     MVX-Net \cite{sindagi2019mvxnet}& LC& 61.0 / 66.1 & 26.0 / 47.8 & 17.6 / 43.1 & 0 / 0 & 59.7 / 54.1 & 34.5 / 30.8 & 28.8 / 25.6 & 0 / 0   \\
     TransFusion \cite{bai2022transfusion}& LC & 66.9 / 70.9 & 29.3 / 51.4 & 20.3 / 45.8 & 0 / 0 & 66.7 / 64.1 & 36.8 / 35.3 & 30.2 / 29.0 & 0 / 0  \\
    
    \bottomrule
   \end{tabularx}
\end{table}

\subsection{A complete analysis of each noisy data case}
\label{subsec:robustness analysis}
Here, we analyze the robustness of existing popular fusion methods on each noisy case proposed in \cref{subsec:benchmark}. \par

\subsubsection{Noisy LiDAR Data}
\paragraph{Limited LiDAR field-of-view.} We investigate the situations when the LiDAR points with limited field-of-view in angle range ($-\pi/3$, $\pi/3$), ($-\pi/2$, $\pi/2$) and ($-0$, $0$). 
The angle range of ($-0$, $0$) is an extreme case when the LiDAR sensor is completely damaged. 
The results are shown in \cref{tab:lidar-failure}. 
For both LiDAR-only and fusion methods, their performance decreases largely in three situations. 
Especially, in the extreme case where all LiDAR points are missing, current fusion methods fail to predict any objects like the LiDAR-only method. 
Thus, for existing fusion methods, the LiDAR modality is the main modality and the camera modality is auxiliary. 
The prediction results of existing fusion methods largely rely on LiDAR information.
There is considerable room for improvement on fusion robustness.
An ideal fusion model should still work as long as there is single modality input.

\begin{table}[h]
    \scriptsize \centering
     \caption{\textbf{Results of the LiDAR object failure case}.}
     \label{tab:object failure}
     \centering
     \vspace{-0.2cm}
     \begin{tabularx}{0.78\textwidth}{l|c|cc|cc}
       \toprule
       \multicolumn{1}{c|}{\multirow{2.8}{*}{Approach}} & \multirow{2.8}{*}{Modality}& \multicolumn{2}{c|}{\robustnuScenes (mAP / NDS)}  & \multicolumn{2}{c}{\robustwaymo  (L2 mAP / L2 mAPH)} \\  
       \cmidrule(r){3-6}
        && clean & object failure & clean & object failure
     \\
     
       \midrule
         CenterPoint \cite{yin2021center} &L& 56.8 / 65.0 &  28.4 / 48.5 & 66.0 / 63.4  & 32.1 / 30.9   \\
        PointAugmenting \cite{wang2021pointaugmenting}&LC& 46.9 / 55.6 & 21.3 / 39.4 &  52.5 / 50.7 & 26.2 / 25.3  \\
        MVX-Net \cite{sindagi2019mvxnet}&LC& 61.0 / 66.1 & 34.0 / 51.1 & 59.7 / 54.1 & 28.7 / 26.0   \\
        TransFusion \cite{bai2022transfusion}&LC & 66.9 / 70.9 & 34.6 / 53.6 & 66.7 / 64.1 &  32.7 / 31.3 \\
       
       \bottomrule
      \end{tabularx}
      \vspace{0.2cm}
    \scriptsize\centering\addtolength{\tabcolsep}{-3.9pt}
     \caption{\textbf{Results of the missing camera inputs case}. \{X\} denotes the location of the missing camera, while the last column indicates the case only keeping the input from the front camera. 
     Note that there is no back camera in the Waymo Open Dataset.}
     \label{tab:missing-camera}
     \vspace{-0.2cm}
     \centering
     \begin{tabularx}{0.94\textwidth}{l|c|cccccccc}
       \toprule
       \multicolumn{1}{c|}{\multirow{2.8}{*}{Approach}} & \multirow{2.8}{*}{~Modality~} & \multicolumn{8}{c}{\robustnuScenes (mAP / NDS)} \\  
       \cmidrule(r){3-10}
        && clean & \{F\} & \{B\} & \{FL\} & \{FR\} & \{BL\} & \{BR\} & Keeping F \\
     
       \midrule
        DETR3D \cite{wang2022detr3d} & C & 34.9 / 43.4 & 25.8 / 39.2 & 23.9 / 38.0 & 28.9 / 39.5 & 29.1 / 39.8 & 30.0 / 40.7 & 29.7 / 40.2 & 3.3 / 20.5   \\
        PointAugmenting \cite{wang2021pointaugmenting}&LC& 46.9 / 55.6 & 42.4 / 53.0 & 41.3 / 52.5 & 43.6 / 53.8 & 45.8 / 54.6 & 45.2 / 54.7 & 44.9 / 54.6 & 31.6 / 46.5 \\
        MVX-Net \cite{sindagi2019mvxnet}&LC& 61.0 / 66.1 & 47.8 / 59.4 & 45.8 / 58.4 & 53.6 / 61.9 & 54.1 / 62.5 & 55.2 / 63.1 & 54.6 / 62.6 & 17.5 / 41.7  \\
        TransFusion \cite{bai2022transfusion}&LC & 66.9 / 70.9 & 65.3 / 70.1 & 66.0 / 70.4 & 66.2 / 70.4 & 66.4 / 70.5 & 66.3 / 70.5 & 66.3 / 70.5 & 64.4 / 69.3  \\
        
        \midrule
        \multicolumn{1}{c|}{\multirow{2.8}{*}{Approach}} & \multirow{2.8}{*}{Modality} & \multicolumn{8}{c}{\robustwaymo  (L2 mAP / L2 mAPH)} \\ 
        \cmidrule(r){3-10}
        && clean & \{F\} & \{B\} & \{FL\} & \{FR\} & \{BL\} & \{BR\} & Keeping F \\
     
       \midrule
        DETR3D \cite{wang2022detr3d} &C& 16.2 / 15.7  & 9.2 / 8.8  & - &13.4 / 13.0  &14.2 / 13.8  & 14.0 / 13.6 & 14.4 / 14.0 & 7.7 / 7.5    \\
        PointAugmenting \cite{wang2021pointaugmenting}&LC& 52.5 / 50.7  & 50.6 / 48.9 & - & 51.8 / 50.0 & 52.1 / 50.3  & 51.8 / 50.0 & 51.9 / 50.1 & 50.2 / 48.4\\
        MVX-Net \cite{sindagi2019mvxnet}&LC& 59.7 / 54.1 & 57.1 / 51.7 & -& 57.5 / 52.2 & 58.1 / 52.7 & 58.5 / 53.1 & 58.9 / 53.5 & 54.3 / 49.2 \\
        TransFusion \cite{bai2022transfusion}&LC & 66.7 / 64.1  & 66.3 / 63.7  & - & 66.5 / 64.0 & 66.4 / 63.8  & 66.4 / 63.8  & 66.5 / 63.9 & 65.8 / 63.2   \\
       \bottomrule
       \multicolumn{10}{l}{{Camera location abbr. F: front. B: back. FL: front-left. FR: front-right. BL: back-left. BR: back-right.} }
      \end{tabularx}
   \end{table}

\paragraph{LiDAR object failure.}
The results of the LiDAR object failure case are shown in \cref{tab:object failure}.
We can find that, with 50\% probability to drop all points of the objects, the performance of both LiDAR-only and LiDAR-camera fusion methods reduce by half approximately.
This indicates current fusion methods fail to work when the foreground LiDAR points are missing, even the objects appear in the images. 
From another perspective, it shows that, for the fusion mechanisms of current LiDAR-camera fusion methods, camera information is not well exploited. 
The fusion process still largely relies on LiDAR information.

\subsubsection{Noisy Camera Sensor}
\paragraph{Missing of camera inputs.}
In the case of missing camera inputs, we consider several combinations of cameras installed in different positions
and report the results in \cref{tab:missing-camera}, in which we can find that the missing front camera or back camera (for nuScenes) has a greater impact on the detection results. 
When all cameras except the front camera are missing, the performance of PointAugmenting and TransFusion decreases no more than 50\% on both \robustnuScenes and \robustwaymo. 
This demonstrates that the robustness of PointAugmenting and TransFusion against camera noise is much better than the other two methods. 
Besides, the performance degradation on \robustwaymo{} is much smaller than that on \robustnuScenes, which indicates the robustness on the various datasets is different.\par

\paragraph{Occlusion of camera lens.}
The results for the case of the dirty camera lens are shown in \cref{tab:mask-camera}.
We surprisingly find all methods have inferior performances when compared to the previous missing camera inputs experiments. It seems the modern deep learning method is akin to a black image instead of occluded ones. This shows a potential new research direction to design a better image feature extractor to address the robustness issue.

\begin{table}[t]
 \scriptsize \centering
  \caption{\textbf{Results of the camera occlusion case}.}
  \label{tab:mask-camera}
  \centering
  \begin{tabularx}{0.76\textwidth}{l|c|cc|cc}
    \toprule
    \multicolumn{1}{c|}{\multirow{2.8}{*}{Approach}} & \multirow{2.8}{*}{Modality}& \multicolumn{2}{c|}{\robustnuScenes (mAP / NDS)}  & \multicolumn{2}{c}{\robustwaymo  (L2 mAP / L2 mAPH)} \\  
    \cmidrule(r){3-6}
     && clean & occlusion & clean & occlusion
  \\
  
    \midrule
     DETR3D \cite{wang2022detr3d} &C& 34.9 / 43.4 & 14.3 / 29.0 & 16.2 / 15.7 &  10.9 / 10.5  \\
     PointAugmenting \cite{wang2021pointaugmenting}&LC& 46.9 / 55.6 & 40.7 / 52.2 & 52.5 / 50.7   &  50.3 / 48.6 \\
     MVX-Net \cite{sindagi2019mvxnet}&LC& 61.0 / 66.1 & 45.5 / 57.6 & 59.7 / 54.1 & 56.4 / 51.1   \\
     TransFusion \cite{bai2022transfusion}&LC & 66.9 / 70.9 & 65.5 / 70.0 & 66.7 / 64.1 &  66.2 / 63.6 \\
    
    \bottomrule
   \end{tabularx}
\end{table}

\begin{figure}[!b]
	\centering
	\begin{minipage}{0.9\linewidth}
        \centerline{\includegraphics[width=\textwidth]{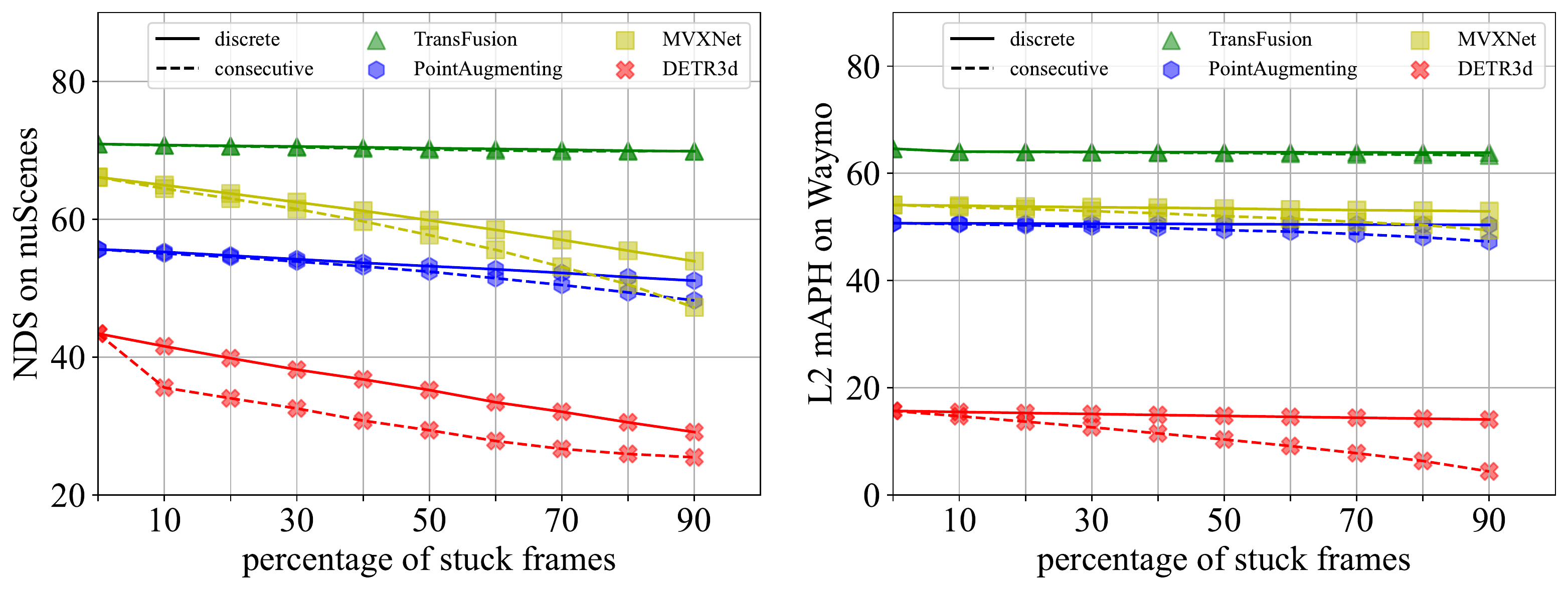}}
        \centerline{(a) Stuck camera frame}
    \end{minipage}
    
    \begin{minipage}{0.9\linewidth}
        \centerline{\includegraphics[width=\textwidth]{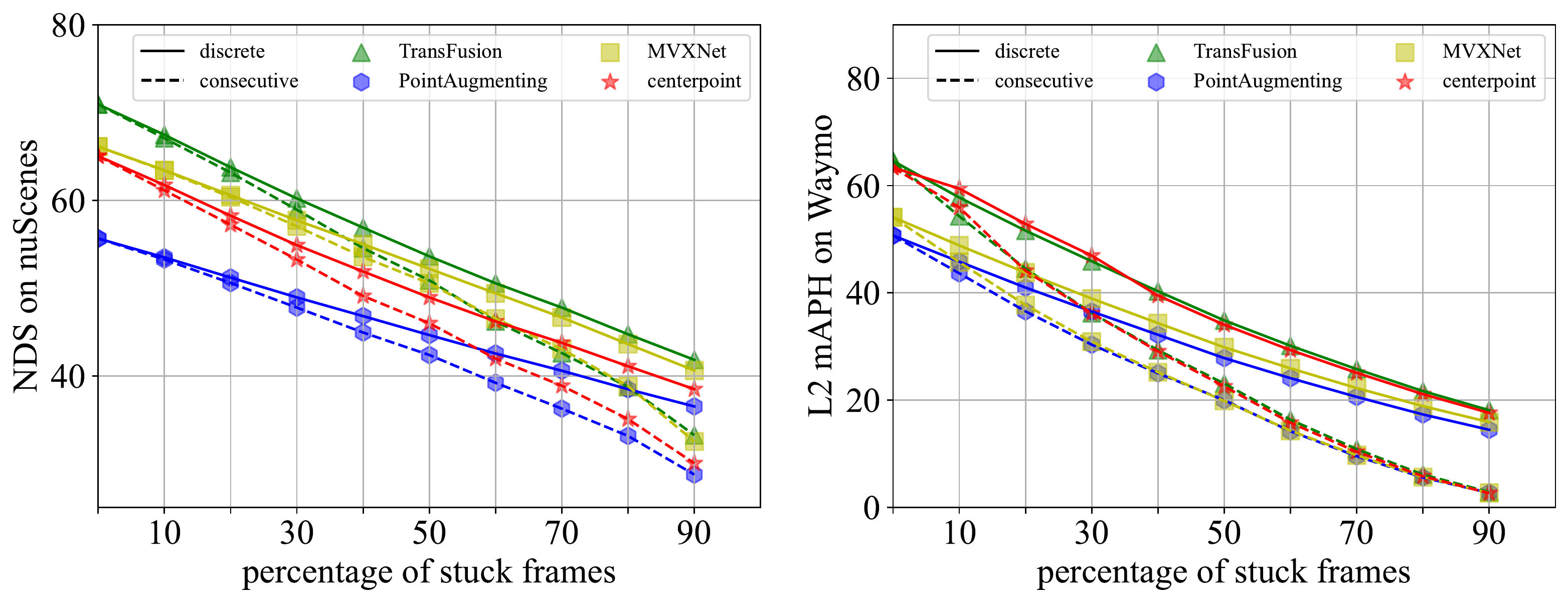}}
        \centerline{(b) Stuck LiDAR frame}
    \end{minipage}
    \caption{\textbf{Temporal misalignment case}. The solid line denotes the discrete selection. The dash line denotes the consecutive selection.}
	\label{fig:temporal}
\end{figure}
\subsubsection{Ill-synchronization}

\paragraph{Spatial misalignment.}
For spatial misalignment, the effect of the noise rotation and translation matrix on fusion models is comparable to that of the camera sensor failure case, as shown in \cref{tab:spatial}. We find that the TransFusion is the most robust cases that on average suffers less than 1pp. compared to the clean settings, where the DETR3D is the most sensitive to the spatial misalignment. \par

\begin{table}[t]
 \scriptsize \centering
 \addtolength{\tabcolsep}{-1.3pt}
  \caption{\textbf{Results of spatial misalignment cases}.}
  \label{tab:spatial}
  \centering
  \begin{tabularx}{0.75\textwidth}{l|c|cc|cc}
    \toprule
    \multicolumn{1}{c|}{\multirow{2.8}{*}{Approach}} &\multirow{2.8}{*}{Modality}& \multicolumn{2}{c|}{\robustnuScenes (mAP / NDS)}  & \multicolumn{2}{c}{\robustwaymo  (L2 mAP / L2 mAPH)} \\  
    \cmidrule(r){3-6}
     && clean & misalignment & clean & misalignment
  \\
  
    \midrule
     DETR3D \cite{wang2022detr3d}&C & 34.9 / 43.4 & 24.2 / 35.0 & 16.2 / 15.7 & 8.0 / 7.8 \\
     PointAugmenting \cite{wang2021pointaugmenting}&LC& 46.9 / 55.6 & 43.6 / 53.8 & 52.5 / 50.7 & 49.8 / 48.1  \\
     MVX-Net \cite{sindagi2019mvxnet}&LC& 61.0 / 66.1 & 50.8 / 59.9 & 59.7 / 54.1 & 54.9 / 49.6   \\
     TransFusion \cite{bai2022transfusion}&LC & 66.9 / 70.9 & 66.5 / 70.7 & 66.7 / 64.1 & 66.3 / 63.7  \\
    
    \bottomrule
   \end{tabularx}
\end{table}

\paragraph{Temporal misalignment.}
For temporal misalignment, we explore 9 levels of severity, in which the proportion of stuck frames is from 10\% to 90\% among all frames in a regular step. 
The results are shown in \cref{fig:temporal}. 
A trend can be observed that the performance degradation of all methods is linear to the percentage of stuck frames among all frames in both discrete selection and consecutive selection. Thus, to reduce the cost of the evaluation, we only choose the case where the stuck frames are 50\% of all frames, whose value is approximately equal to the average of 9 levels of severity.
Interestingly, although TransFusion performs well against the stuck camera frame case, we can observe that
the perfromance of TransFusion decreases faster than other fusion methods when the LiDAR stuck frame ratio increases, showing that it is more sensitive than other fusion methods.

\section{Discussion and Future Work}
Though the main contribution of this work is to provide a systematic overview of different aspects of the perception system and construct a robustness benchmark, we nonetheless provide a simple yet effective baseline method, robustness finetuning, to improve the robustness of these methods. 

\subsection{A simple baseline to improve the robustness: Robust Finetuning}

We provide a simple baseline method by treating our toolkit as a data augmentation method to enrich the training data as the first attempt to improve the robustness of performance. Specifically, we use the toolkit to transform the training data. Different from the evaluation, where reproducibility is a must, we can randomly generate corresponding noisy data during training. However, in practice, we discover that transforming all data into a noisy format will significantly decrease the performance when there is no such noise. To this end, we propose a cascaded augmentation policy during training: i) if a randomly sampled float number is higher than an augmentation probability $p_a$, proceed, otherwise use the normal clean data; 
ii) sampling one transformation from all robustness cases according to the probability distribution $p_o$, then proceeding data with the transformation.
See \cref{subsec:robust train setting} for more details.

\begin{table}[h]
  \scriptsize \centering \addtolength{\tabcolsep}{-2.5pt}
  \caption{\textbf{Robust training results}.}
  \label{tab:finetune}
  \centering
 \begin{tabularx}{0.9\textwidth}{l|c|c|cc|cc|cc}
    \toprule
    \multicolumn{1}{c|}{\multirow{2.8}{*}{Approach}} &\multirow{2.8}{*}{Modality}& & \multicolumn{2}{c|}{Overall} & \multicolumn{2}{c|}{Lidar}  & \multicolumn{2}{c}{Camera} \\  
    \cmidrule(r){3-9}
 && $P_{C}$ & $mP_{R}$ & $R$ & $mP_{R}$ & $R$ & $mP_{R}$ & $R$ \\
   \midrule
   \multicolumn{9}{c}{\robustnuScenes (mAP / NDS)}                 \\
    \midrule
     MVX-Net \cite{sindagi2019mvxnet}&LC & 61.0 / 66.1 & 38.4 / 53.6 & 0.63 / 0.81 & 26.4 / 47.3 & 0.43 / 0.72 & 44.3 / 56.7 & 0.73 / 0.86 \\
     MVX-Net + finetune&LC & 59.4 / 65.0 & 40.9 / 54.8 & 0.69 / 0.84 & 24.6 / 46.1 & 0.41 / 0.71 & 49.1 / 59.2 & 0.83 / 0.91 \\
    \midrule
  \multicolumn{9}{c}{\robustwaymo  (L2 mAP / L2 mAPH)}       \\
    \midrule
     MVX-Net \cite{sindagi2019mvxnet}&LC & 59.7 / 54.1 & 44.3 / 40.1 & 0.74 / 0.74 & 28.3 / 25.5 & 0.47 / 0.47 & 56.4 / 51.0 & 0.94 / 0.94 \\ 
     MVX-Net + finetune &LC & 59.5 / 54.0 & 47.7 / 43.1 & 0.80 / 0.80 & 27.8 / 25.2 & 0.47 / 0.47 & 57.6 / 52.1 & 0.97 / 0.96 \\

    \bottomrule
   \end{tabularx}
\end{table}

We select the MVXNet to study the effectiveness of our method, as it has the most balanced LiDAR and camera performance, and report the results in  \cref{tab:finetune}. We observe that, though applying such robustness training slightly deteriorates the performance on the clean dataset, it significantly improves the robustness, where the mean robustness $R$ improves from 0.63 and 0.81 to 0.69 and 0.84 in terms of mAP and NDS on \robustnuScenes{}, and from 0.74 to 0.80 in terms of L2 mAP and mAPH on \robustwaymo{}. However, we can still see a large gap between the robust benchmark and the clean ones, evidencing there is an actual research gap in this research direction. 

\subsection{Future research directions}
In general, we believe an ideal sensor fusion framework should be able to do the following: i) given both modality data, it can significantly surpass the performance of single modality methods; ii) when there is a disruption of one modality, the performance should not be worse than the single modality method of the other. Currently, this approach is handled by using comprehensive post-processing techniques of the perception system. We hope our robust benchmark can be a tool for the community to fully exploit this research direction to develop truly robust methods that can be deployed on realistic vehicles.

\newpage
\small
\bibliographystyle{plain}
\bibliography{ref}

\newpage

\section{Appendix}
\subsection{Autonomous driving datasets}
\label{apdx:dataset} 

\textbf{nuScenes Dataset.} 
nuScenes is a large-scale autonomous-driving dataset for 3D detection, consisting of 700, 150 and 150 scenes for training, validation, and testing, respectively. 
Each frame contains one point cloud and six calibrated images that cover 360 fields-of-view. 
For 3D detection, the main metrics are mean Average Precision (mAP) and nuScenes detection score (NDS). 
The mAP is defined by the BEV center distance with thresholds of {0.5m, 1m, 2m, 4m}, instead of the IoUs of bounding boxes. 
NDS is a consolidated metric of mAP and other metric scores, such as average translation error and average scale error.

\textbf{Waymo Open Dataset.} 
Waymo Open Dataset is an another large-scale dataset for autonomous driving, which contains 798 training, 202 validation, and 150 testing sequences. 
Each sequence has about 200 frames with LiDAR points and camera images, which are collected by five LiDAR sensors and five pinhole cameras.
The official metrics are mean Average Precision (mAP) and mean Average Precision weighted by Heading (mAPH). 
The mAP and mAPH are defined based on the 3D IoU with the threshold of 0.7 for vehicles and 0.5 for pedestrians and cyclists. 
The measures are reported based on the distances from objects to sensor, i.e., 0-30m, 30-50m and >50m, respectively. 
Besides, two difficulty levels, LEVEL 1 (boxes with more than five LiDAR points) and LEVEL 2 (boxes with at least one LiDAR point), are considered.

\subsection{Implementation Details of Fusion Models}
\label{subsec:train setting}

For the sake of better reproducibility, we re-implement the MVX-Net into the MMDetection3D\cite{mmdet3d} framework. Here we detail the implementation settings. 

\paragraph{MVX-Net.}
Following the original paper~\cite{sindagi2019mvxnet}, we use \pp{}~\cite{lang2019pointpillars} as the LiDAR stream and ResNet50\cite{he2016resnet} with FPN~\cite{lin2017FPN} as the image stream. At the fusion stage, we 
project each LiDAR point to all images from different views to acquire the corresponding image features from the deep networks. Then, we average the features on the channel dimension and then concatenate all features from different views with the original LiDAR point feature before the downstream task.
During training, we use the same training schedules and hyper-parameters of \pp{}, where Adam optimizer is used with learning rate 0.001, weight decay 0.01, batch size 16, epoch 12, and learning rate decay to 1/10 at 8 and 11 epochs.

On the \nus{} dataset, we set the detection region of interest to $[-50m, 50m]$ for the X and Y axis, and $[-5m, 3m]$ for the Z axis. The pillar size is kept as $[0.25m, 0.25m]$. The image feature extractor is trained for 36 epochs on nuImage\cite{nuscenes2019}. On \waymo{} dataset, we set the detection range to $[-74.88m, 74.88m]$ for the X and Y axis, and $[-2m, 4m]$ for the Z axis. The pillar size is hold as $[0.32m, 0.32m]$. The image feature extractor is trained for 36 epochs on \waymo{} of the 2D detection task.

\begin{table}[h]
 \tiny\centering\addtolength{\tabcolsep}{-3.3pt}
  \caption{\textbf{Ablation of robustness finetuning}}
  \label{tab:finetune-detail}
  \centering
 \begin{tabularx}{0.98\textwidth}{c|c|ccc|ccc|cccc}
    \toprule
    \multirow{2}{*}{Approach}&\multirow{2}{*}{Aug}&&&& \multicolumn{3}{c|}{Lidar}  & \multicolumn{4}{c}{Camera} \\  
    \cmidrule(r){3-12}
     & & $P_{C}$ & $mP_{R}$ & $R$ & Stuck & FOV & Object & Stuck & Missing & Occlusion  & Calib 
  \\ \midrule
  \multicolumn{11}{c}{\robustnuScenes (mAP / NDS)}                 \\
    \midrule
    \multirow{6}{*}{MVX-Net\cite{sindagi2019mvxnet}} & None & 61.0 / 66.1 & 38.4 / 53.6 & 0.63 / 0.81 & 35.2 / 51.4 & 17.6 / 43.1 & 34.0 / 51.1& 48.3 / 58.8 & 32.7 / 50.6 & 45.5 / 57.6 & 50.8 / 59.9 \\
    & LIDAR stuck &  58.4 / 64.2 &   35.6 / 51.5& 0.61 / 0.80  & 39.1 / 52.1& 17.8 / 42.8& 34.2 / 51.4 & 42.4 / 54.9& 28.2 / 47.6& 41.0 / 54.8& 46.6 / 56.9  \\
    & FOV &  54.2 / 59.1 & 33.9 / 49.9  & 0.63 / 0.84 & 33.3 / 49.5& 18.4 / 44.0& 30.3 / 47.4 & 41.4 / 50.8 & 27.7 / 47.0 & 40.5 / 54.9  & 45.7 / 56.0   \\
    & Object &  56.8 / 60.9 &  36.5 / 51.4  & 0.63 / 0.84 & 39.1 / 52.1& 17.8 / 42.8& 33.1 / 49.9 & 44.3 / 53.7 & 29.9 / 48.5  & 42.8 / 55.1 & 48.2 / 57.5   \\
    & Camera stuck &  58.6 / 64.7 &   38.7 / 53.7& 0.64 / 0.84& 31.0 / 48.9& 17.2 / 42.9& 32.7 / 50.0& 51.3 / 60.4& 35.2 / 51.8& 45.7 / 57.6& 52.0 / 60.9  \\
    & Missing & 60.9 / 66.0&   41.1 / 55.0& 0.66 / 0.83 & 34.9 / 51.1& 18.3 / 43.6& 34.1 / 50.8 & 48.7 / 59.1& 45.1 / 57.1& 48.2 / 58.9& 51.1 / 60.2  \\
    & Calib & 60.0 / 65.4  &   39.4 / 54.0& & 33.6 / 50.3& 17.8 / 43.3& 33.6 / 50.7 & 50.0 / 59.6& 34.6 / 51.3& 45.8 / 57.6& 54.9 / 62.2  \\

    \bottomrule
   \end{tabularx}
\end{table}
\begin{table}[h]
  \centering
  \caption{Probability distribution of transformation selection during finetuning.}
  \label{tab:probability}
  \centering
 \begin{tabularx}{0.86\textwidth}{c|ccc|cccc}
    \toprule
    \multicolumn{1}{c|}{\multirow{2.8}{*}{Transformation}}& \multicolumn{3}{c|}{LiDAR}  & \multicolumn{4}{c}{Camera} \\  
    \cmidrule(r){2-8}
     & Stuck & FOV & Object & Stuck & Missing & Occlusion  & Calib \\
    \midrule
    Probability & 0 & 0 & 0 & 1/3 & 1/3 & 0 & 1/3 \\
    \bottomrule
  \end{tabularx}
\end{table}
\subsection{Ablation study of robustness finetuning}
\label{subsec:robust train setting}

To determine the probability distribution $p_o$ of all transformations, we first analyze the effects of each individual transformation, as shown in \cref{tab:finetune-detail}. 
It's worth noticing that the transformation of camera lens occlusion case is included during finetuning stage, since the masks for simulating the occlusion is only available during validation. 
From \cref{tab:finetune-detail}, we can find that, compared with MVX-Net baseline trained on clean data, finetuning with noisy LiDAR data decreases the average performance $mP_R$ on each noisy fusion cases. 
By contrast, finetuning with noisy camera data improves the average performance $mP_R$.
Thus, the sampling probabilities of noisy LiDAR transformations are set to be zero and the sampling probabilities of remaining noisy camera transformations, i.e., camera-stuck and missing of camera input, are set to be $1/3$.
The final probability distribution $p_o$ is listed in \cref{tab:probability}.

\end{document}